\def\year{2020}\relax
\documentclass[letterpaper]{article} 
\usepackage{aaai20}  
\usepackage{times}  
\usepackage{helvet} 
\usepackage{courier}  
\usepackage[hyphens]{url}  
\usepackage{graphicx} 
\usepackage{amsmath}
\usepackage{amssymb}
\usepackage{amsthm}
\usepackage{booktabs}
\usepackage{multirow}
\usepackage{verbatim}

\usepackage{graphicx}  
\newcommand{\citet}[1]{\citeauthor{#1} \shortcite{#1}}
\newcommand{\citep}{\cite}

\title{A Critique of the Smooth Inverse Frequency Sentence Embeddings}
\author{Aidana Karipbayeva\\College of Liberal Arts and Sciences\\University of Illinois at Urbana-Champaign\\aidana2@illinois.edu\And Alena Sorokina\\School of Sciences and Humanities\\Nazarbayev University\\alena.sorokina@nu.edu.kz\And Zhenisbek Assylbekov\\School of Sciences and Humanities\\Nazarbayev University\\zhassylbekov@nu.edu.kz
}
 
\begin{document}

\maketitle

\begin{abstract}
We critically review the smooth inverse frequency sentence embedding method of \citeauthor{arora2017simple} \shortcite{arora2017simple}, and show inconsistencies in its setup, derivation and evaluation.
\end{abstract}

\section{Introduction}
The smooth inverse frequency (SIF) sentence embedding method of \citeauthor{arora2017simple} \shortcite{arora2017simple} has gained attention in the NLP community due to its simplicity and competetive performance. We recognize the strengths of this method, but we argue that its theoretical justification contains a number of flaws. In what follows we show that there are contradictory arguments in the setup, derivation and experimental evaluation of SIF. 

We first recall the word production model used by the authors as the foundation of SIF: given the context vector $\mathbf{c}\in\mathbb{R}^d$, the probability that a word $w$ is emitted in the context is modeled by
\begin{align}
    p(w\mid\mathbf{c})=\alpha p(w)+(1-\alpha)\frac{\exp(\langle\mathbf{w}, \tilde{\mathbf{c}}\rangle)}{Z_{\tilde{c}}}\label{eq:gen_model}\\
    \text{with}\,\tilde{\mathbf{c}}=\beta\mathbf{c}_0+(1-\beta)\mathbf{c},\quad\mathbf{c}_0\perp\mathbf{c},
\end{align}
where $\alpha, \beta\in[0,1]$ are scalar hyperparameters, $\mathbf{w}\in\mathbb{R}^d$ is a word embedding for $w$, $\mathbf{c}_0\in\mathbb{R}^d$ is the so-called common discourse, and $Z_{\tilde{\mathbf{c}}}=\sum_{w\in\mathcal{W}}\exp(\langle\tilde{\mathbf{c}},\mathbf{w}\rangle)$ is the normalizing constant.

\section{Inconsistent Setup}
The authors empirically find (see their section 4.1.1) that the optimal value of $\alpha$ satisfies
\begin{equation}
    10^{-4}\le\frac{1-\alpha}{\alpha Z}\le10^{-3},\label{eq:a_bound}
\end{equation}
where $Z=\mathbb{E}[Z_{\mathbf{c}}]$. In their previous work, \citeauthor{arora2016latent} \shortcite{arora2016latent} showed (see the proof sketch of their Lemma 2.1 on p.~398) that under isotropic assumption on $\mathbf{w}$'s,
\begin{equation}
    \mathbb{E}_\mathbf{w}[Z_\mathbf{c}]=n\mathbb{E}_\xi[\exp(\xi^2\|\mathbf{c}\|^2/2)]\label{eq:expectation_z_c}
\end{equation}
where $\xi$ is a random variable upper bounded by a constant. From \eqref{eq:expectation_z_c} we have $n\le Z=\mathbb{E}[Z_\mathbf{c}]$, and combining this with the right inequality from \eqref{eq:a_bound} we have $
    \frac{1-\alpha}{\alpha}\le\frac{1}{10^3 n}    \,\Leftrightarrow\,\frac{10^3n}{10^3n+1}\le\alpha$. For a typical vocabulary size $n=10^5$ this implies $0.99999999\le\alpha\le1$, which means that the generative model \eqref{eq:gen_model} is essentially a unigram model $\Pr(w\mid\mathbf{c})\approx p(w)$ that practically ignores the context. 

\section{Contradictory Derivation}
Treating any sentence $s$ as a sequence of words $[w]$, the authors construct its log-likelihood given the smoothed context vector $\tilde{\mathbf{c}}$ as $
\ell(\mathbf{c})=\sum_{w\in s}\log\Pr(w\mid\mathbf{c})$. Then this log-likelihood is linearized using Taylor expansion at $\tilde{\mathbf{c}}=\mathbf{0}$:
\begin{equation}
\ell(\tilde{\mathbf{c}})\approx\ell(\mathbf{0})+\nabla\ell(\mathbf{0})^\top\tilde{\mathbf{c}},\label{ell_linear}
\end{equation}
and after that the right-hand side of \eqref{ell_linear} is optimized with $\tilde{\mathbf{c}}$ constrained to take values on the unit sphere $\{\tilde{\mathbf{c}}\in\mathbb{R}^d\mid\|\tilde{\mathbf{c}}\|=1\}$, which contradicts the assumption $\tilde{\mathbf{c}}\approx\mathbf{0}$ needed for the linear approximation  \eqref{ell_linear} to be adequate.

\begin{figure*}
    \centering
    \includegraphics[width=.49\textwidth]{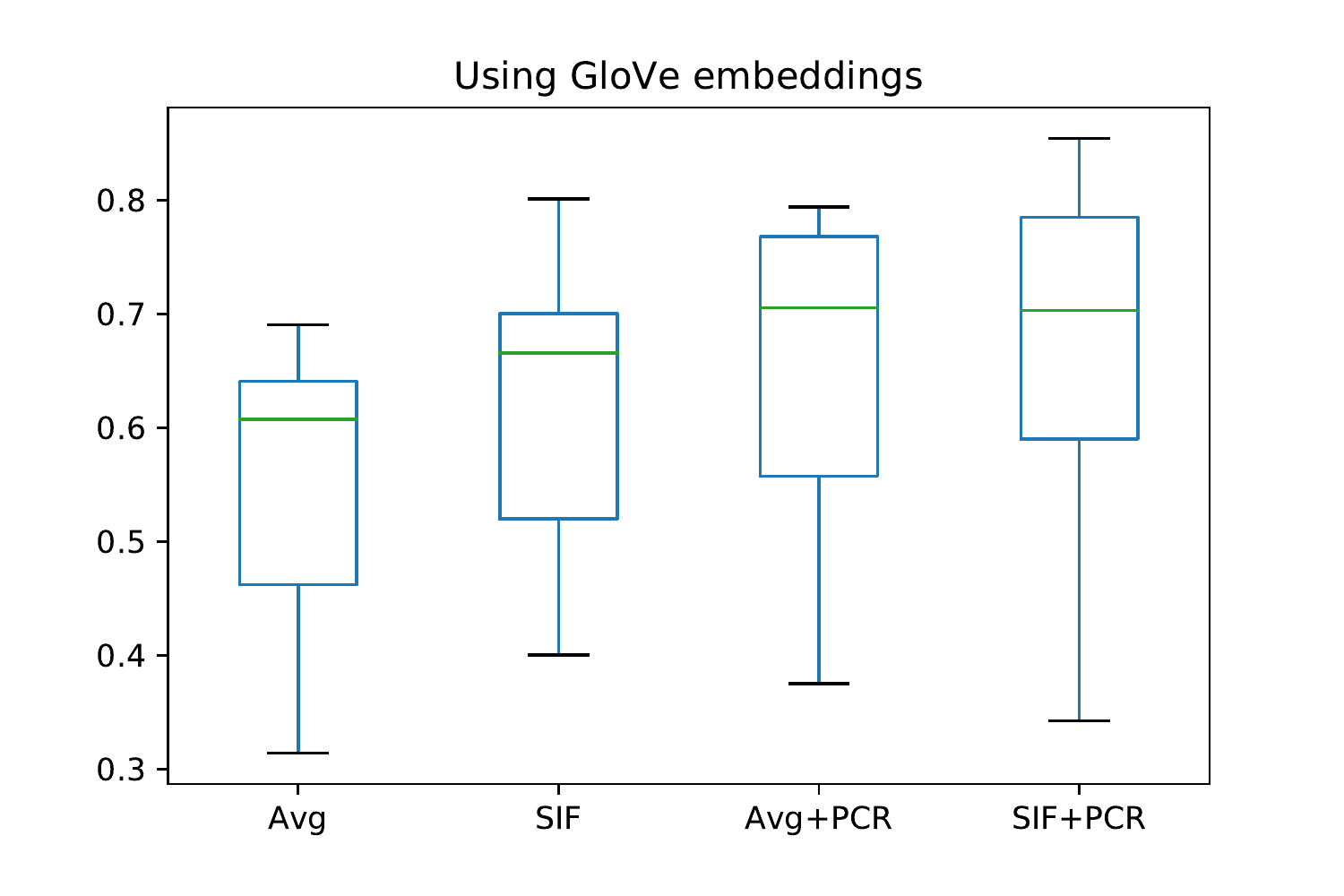}\includegraphics[width=.49\textwidth]{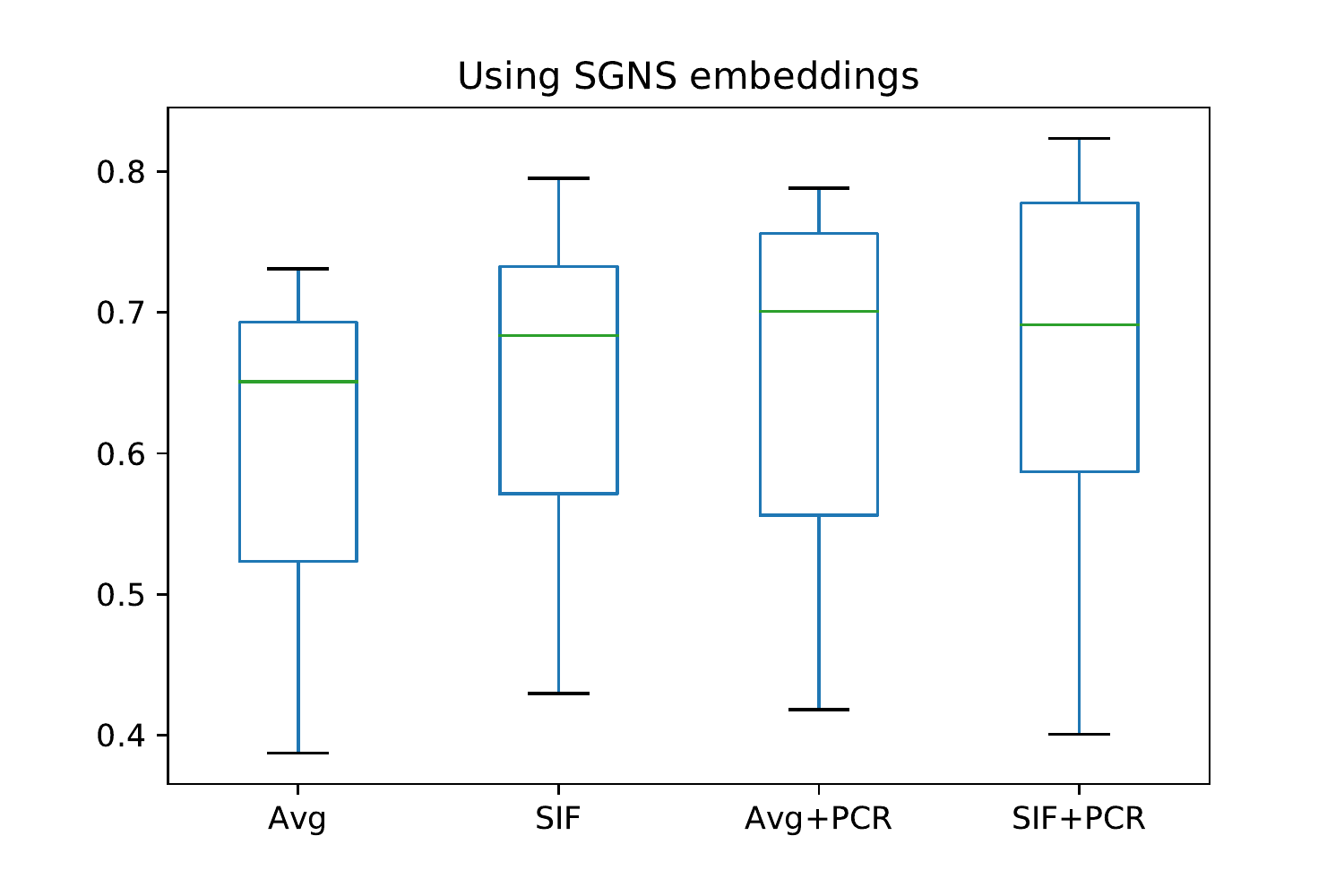}
    \caption{Performance of SIF vs Simple Average on STS tasks. The evaluation criterion is the Pearson’s coefficient between the predicted scores and the ground-truth scores.}
    \label{fig:sif_vs_avg}
\end{figure*}

\section{Model Inadequacy}
It \textit{is} possible to have a valid derivation of the SIF sentence embedding as the Maximum-a-Posteriori (MAP) estimate of $\mathbf{c}$ given $s$ once we assume the generative model 
\begin{equation}\textstyle
    p(w\mid \mathbf{c}) \propto \exp\left(\frac{a}{{p_i} + a}\langle\mathbf{w}, \tilde{\mathbf{c}}\rangle\right)\label{eq:new_gen_model}
\end{equation}
instead of \eqref{eq:gen_model}. The proof is similar to that of Lemma~3.1 in \citeauthor{arora2016latent} \shortcite{arora2016latent}, and is left as exercise. Now, assume that $c$ is a single context word, and $\mathbf{c}$ is its embedding. Taking logarithm of both sides in \eqref{eq:new_gen_model}, then solving for $\langle\mathbf{w},\tilde{\mathbf{c}}\rangle$ and assuming that the normalizer in \eqref{eq:new_gen_model} concentrates well around a constant $Z$, we have
\begin{equation}\textstyle
\langle\mathbf{w}, \tilde{\mathbf{c}}\rangle \approx \underbrace{\frac{{p(w)} + a}{a}\left(\log p(w\mid c)+\log Z\right)}_{\mathbf{M}_{w,c}}\label{eq:factorization}
\end{equation}
This means that the word and context embeddings that underlie the language model \eqref{eq:new_gen_model} give a low-rank approximation of a matrix $\mathbf{M}$ in which the element in row $w$ and column $c$ is equal to the right-hand side of \eqref{eq:factorization}. It is well known that the word and context embeddings that underlie the SGNS training objective give a low-rank approximation of the shifted PMI matrix, and that factorizing the latter with truncated SVD gives embeddings of similar quality \cite{levy2014neural}. This means, that if the model \eqref{eq:new_gen_model} is adequate, then the truncated SVD of $\mathbf{M}$ should give us good-quality word embeddings as well. We calculated the shifted PMI and $\mathbf{M}$ on \texttt{text8} data\footnote{\url{http://mattmahoney.net/dc/textdata.html}} using vocabulary size $35000$ and then performed rank-200 approximation. The resulting embeddings were evaluated on standard similarity (WordSim) and analogy (Google and MSR) tasks. The hyperparameter $Z$ was tuned using grid search and the optimal value was s.t. $\log Z=13$. The results of evaluation are given in Table~\ref{tab1}.%
\setlength{\tabcolsep}{8pt}
\begin{table}
\begin{center}
\begin{tabular}{c c c c c}
\toprule
Matrix & Similarity task & Analogy Task \\
\midrule
$\mathbf{M}$ & 61.18 & 27.14\\
PMI & 65.05 & 29.58 \\
\bottomrule
\end{tabular}
\caption{Word embeddings from PMI and $\mathbf{M}$. For word similarities evaluation metric is the Spearman's correlation with the human ratings, while for word analogies it is the percentage of correct answers.}
\label{tab1}
\end{center}
\end{table}
As we can see the word embeddings that underlie \eqref{eq:new_gen_model} do not outperform those that underlie SGNS. Hence, the adequacy of \eqref{eq:new_gen_model} is questionable.

\section{Evaluation Flaws}
In fact, the method of \citeauthor{arora2017simple} \shortcite{arora2017simple} is not only in using SIF weights but also in removing the principal component from the resulting sentence embeddings. When the authors evaluate their method against a simple average (Avg) of word vectors, they do not consider principal component removal (PCR) as a separate factor, i.e. they do not compare against a simple average of word embeddings followed by a principal component removal (Avg+PCR). We performed such comparison on datasets from the SemEval Semantic Textual Similarity (STS) tasks\footnote{\url{http://ixa2.si.ehu.es/stswiki/index.php/Main_Page}} with {\sc GloVe} and {\sc SGNS} embeddings, and the results are illustrated on Fig.~\ref{fig:sif_vs_avg}. As we can see, SIF is indeed stronger than Avg, but this advantage is diminished when we remove the principal components from both. Looking at the boxplots, one may think that the difference between Avg+PCR and SIF+PCR is not significant, however this is not the case: SIF+PCR demonstrates higher scores than Avg+PCR according to paired one-sided Wilcoxon signed rank test, with p-values $<0.02$ for both GloVe and SGNS embeddings. Thus, we admit that the overall claim of the authors is valid: SIF outperforms Avg with and without PCR.

\section{Conclusion}
The sentence embedding method of \citeauthor{arora2017simple} \shortcite{arora2017simple} is indeed a simple but tough-to-beat baseline, which has a clear underlying intuition that the embeddings of too frequent words should be downweighted when summed with those of less frequent ones. However, one does not need to tweak a previously developped mathematical theory to justify this empirical finding: in pursuit of mathematical validity, the SIF authors made a number of errors.

\bibliography{ref}
\end{document}